\documentclass[10pt]{article}

\usepackage[preprint]{tmlr}

\usepackage{graphicx}
\usepackage{hyperref}
\usepackage{booktabs}       
\usepackage{amsfonts}       
\usepackage{amsmath}        
\usepackage{nicefrac}       
\usepackage{microtype}      
\usepackage{xcolor}         
\usepackage{cleveref}      
\usepackage{amsmath}
\usepackage{graphicx}
\usepackage{subcaption} 
\usepackage{algorithm}
\usepackage{algorithmic}
\usepackage{etoolbox}
\makeatletter
\let\TMLR@AND\AND
\BeforeBeginEnvironment{algorithmic}{\let\AND\@undefined}
\AfterEndEnvironment{algorithmic}{\let\AND\TMLR@AND}
\makeatother
\usepackage{multirow}

\usepackage{orcidlink}

\newcommand{\update}[1]{{\textcolor{black}{#1}}}

\newcommand{\method}[0]{NoiseShift}

\begin{document}

\title{NoiseShift: Resolution-Aware Noise Recalibration for Low-Resolution Image Generation}

\author{\name Ruozhen He \quad Moayed Haji-Ali \quad Ziyan Yang \quad Vicente Ordonez \\
      \addr Rice University \\
      \addr \{catherine.he, mh155, zy47, vicenteor\}@rice.edu}

\maketitle

\begin{abstract}
Text-to-image diffusion models often degrade when sampled at resolutions outside the final training resolution set.
\update{Prior work has largely emphasized higher resolution generation, enabling pretrained diffusion models to extrapolate beyond the resolutions seen during training. In this work, we instead target lower-resolution generation, performing inference at reduced resolution to significantly cut computational cost.}
\update{We show that network \textit{conditioning} of the noise level induces a train--test mismatch that directly degrades low-resolution generation: the same scheduled noise level can correspond to a different perceptual corruption level at lower resolutions, mis-calibrating the denoiser timestep and noise embedding.}
\update{To this end, we propose \method{}, a training-free recalibration method that keeps the original \textit{noise sampling schedule} unchanged and instead re-indexes the \textit{noise conditioning} of the denoiser to restore local forward–reverse consistency.
Using a lightweight coarse-to-fine calibration on a small set of image--text pairs, \method{} learns a resolution-specific mapping from scheduler noise to conditioning noise, reducing train-test mismatch and improving lower-resolution generation quality.
}
When \method{} is applied to Stable Diffusion 3 (SD3), Stable Diffusion 3.5 (SD3.5), and Flux-Dev, generation quality at low resolutions improves consistently. Particularly, SD3 generation at 128$\times$128 resolution gets an improved FID score from 203 to 171, and SD3.5 gets an improved FID score from 310 to 277 on LAION-COCO. Even Flux-Dev which already implements a complementary time-shifting strategy gets a modest boost from \method{} with an improved FID score from 120 to 113 at 64$\times$64 resolution. 
\update{More importantly, \method{} achieves such improvements with minimal implementation changes and no additional inference overhead.}
\end{abstract}

\section{Introduction}
\label{sec:intro}

\begin{figure}[t]
    \centering
    \includegraphics[width=\linewidth]{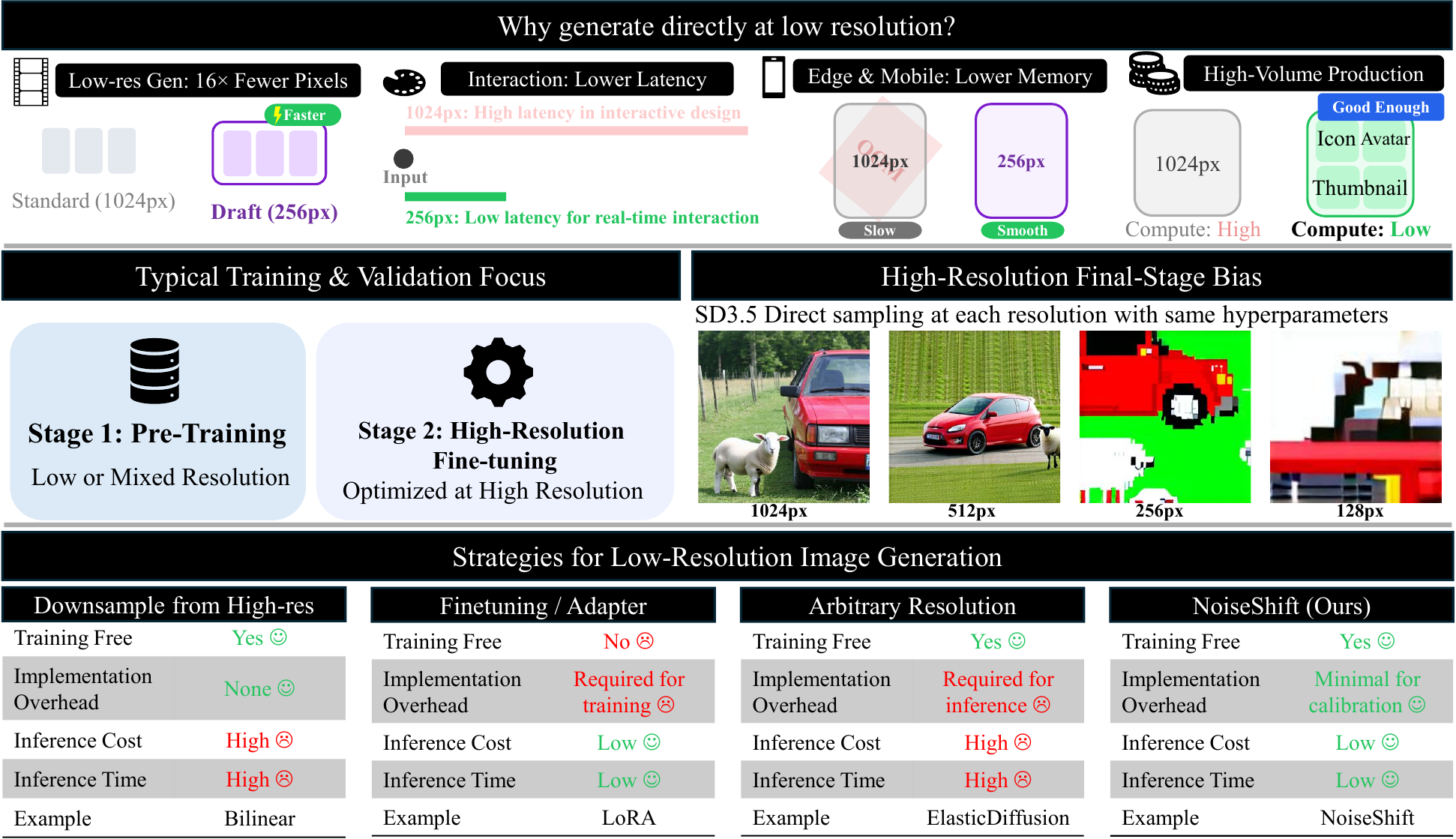}
\caption{\textbf{Motivation for direct low-resolution sampling.}
Direct low-resolution generation enables fast generation and interaction, reduced memory on edge/mobile devices, and higher-throughput production.
However, modern diffusion pipelines are typically optimized and validated at a final high resolution, which can bias them toward high-resolution synthesis and degrade quality when sampled directly at lower resolutions (example shown for SD3.5~\cite{Esser2024ScalingRF} with the same prompt and sampling hyperparameters).
The bottom panel further situates \method{} relative to common alternatives, including downsampling, finetuning/adapters, and arbitrary-resolution inference methods.}
    \label{fig:motivation}
\end{figure}

Large text-to-image diffusion models~\cite{peebles2023scalable,rombach2022high,sdxl, dfm} are typically optimized and validated at a target resolution~\cite{shen2024diffclip,zhang2023adding,xiao2024omnigen}, yet practical deployment often requires \emph{resolution flexibility}. Direct low-resolution sampling is valuable in several common settings: creators rely on inexpensive generated content before committing to full-resolution renders, large-scale systems generate many candidates for filtering or ranking, and interactive or edge applications benefit from lower latency and memory usage~\cite{huang2015automatic, kolotouros2024instant}. In these scenarios, generating directly at lower resolution can substantially reduce \update{sampling latency,}
enabling faster iteration and higher throughput. At the same time, such benefits are only attractive if adaptation remains lightweight in practice. Ideally, a method should preserve the original sampling pipeline, avoid model retraining or architectural changes, and introduce little to no additional inference overhead.

However, directly sampling a pretrained model at resolutions below its final training resolution often leads to severe degradation, including artifacts, unstable structure, and inconsistent semantics.
This issue is widely observed~\cite{peebles2023scalable,unet}. Many modern pipelines emphasize a final high-resolution stage where model selection and hyperparameter choices implicitly prioritize that final target resolution.
The resulting \emph{high-resolution final-stage bias} can deliver excellent quality at $1024\!\times\!1024$ while leaving the model poorly calibrated for low-resolution sampling, as qualitatively shown in~\Cref{fig:motivation}.
Existing solutions typically sidestep the problem by (a) Generating first at high resolution and then downsampling, which sacrifices efficiency, (b) Retraining or using Adapter finetuning~\cite{sdxl,Cheng2024ResAdapterDC} which is costly for modern large diffusion models~\cite{Esser2024ScalingRF, flux2024}, or (c) Introducing inference-time changes for arbitrary-size or multi-stage generation~\cite{elasticdiffusion, multidiffusion, foleycrafter, any-size-diffusion, freescale, fouriscale,demofusion}.
\update{The closest to our work are Schedule-based methods such as time-shifting~\cite{timeshifting}, which adjust the \emph{sampling schedule} to better match different resolutions and mitigate exposure bias. These methods have proven effective on large-scale diffusion models~\cite{flux2024}. However, by recalibrating the denoiser noise conditioning rather than changing the sampling schedule, \method{} addresses a complementary source of mismatch and can further reduce exposure bias, yielding additional gains in generation quality.  }

In this work, we focus on a complementary and practical source of cross-resolution degradation: \emph{noise-level conditioning mismatch}.
In diffusion and flow-matching samplers, the noise schedule $\{\sigma_t\}$ determines the reverse integration trajectory. \update{Critically, at each sampling step, the denoiser network relies on explicit conditioning on $\sigma_t$ to interpret the level of input corruption. However, when sampling at unseen lower resolutions, the same scheduled noise level can induce a different effective corruption pattern in the input, making the default conditioning misaligned with the true perceptual corruption level.}

We propose \method{}, a training-free test-time calibration method that corrects this mismatch by \emph{re-indexing the denoiser's noise-level conditioning} while keeping the sampling schedule unchanged.
Concretely, we preserve the original schedule $\{\sigma_t\}$ (thus preserving the sampling trajectory) and learn a resolution-specific conditioning sequence $\{\hat{\sigma}_t\}$ used \emph{only} for the denoiser's timestep/noise embedding---i.e., we keep what the sampler does fixed, but change what the network is told about the noise level. This distinction is important because it is separate from changing the sampling schedule itself~\cite{exposurebias,timeshifting,resolutionchromatography}: even if the sampler follows the original $\{\sigma_t\}$, the model benefits from receiving a different noise-level input 
\update{that better reflects the effective corruption level of the input.}

\begin{figure}[t]
    \centering
    \includegraphics[width=0.95\linewidth]{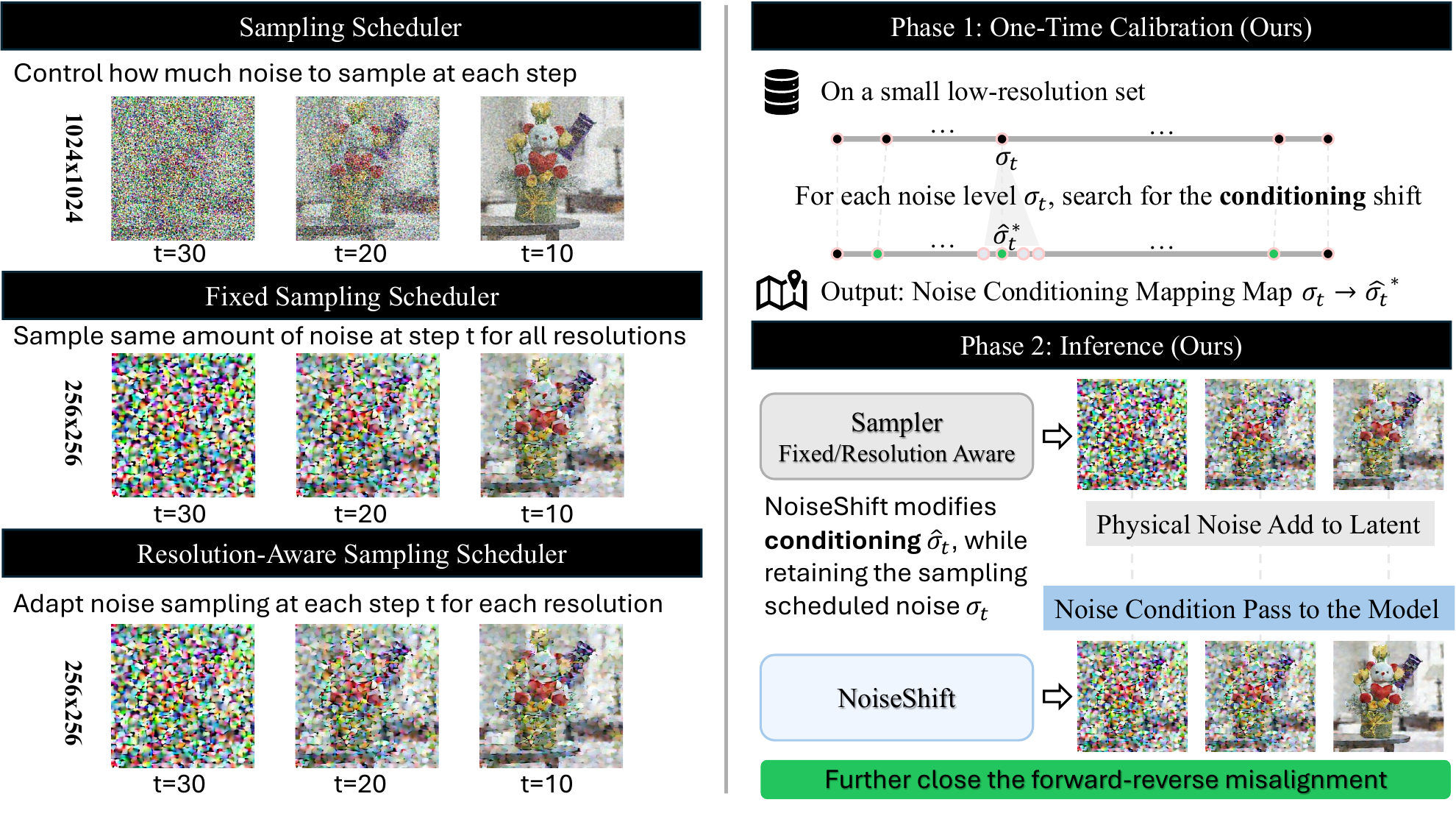}
\caption{\textbf{\method{} overview: calibrating noise-level conditioning without changing the sampling schedule.}
The sampling scheduler controls the scheduled noise level $\sigma_t$ (fixed across resolutions or resolution-aware), which governs the physical noise added during sampling.
\method{} performs a one-time per-resolution calibration on a small low-resolution set to learn a mapping to a calibrated conditioning noise $\hat{\sigma}_t^*$, and then reuses this mapping at inference by conditioning the denoiser on $\hat{\sigma}_t^*$ while keeping the scheduled noise $\sigma_t$ unchanged, enabling compatibility with both fixed and resolution-aware schedulers. 
}
    \label{fig:pipeline}
\end{figure}

As shown in \Cref{fig:pipeline}, \method{} performs a lightweight coarse-to-fine calibration using a small set of image--text pairs, producing a mapping from scheduled noise to conditioning noise that can be cached once per resolution and reused for unseen prompts and images at inference.

This design makes \method{} both minimal and composable.
As it operates on conditioning rather than on the sampling schedule, \method{} is compatible with schedule-based approaches such as time shifting.
As a result, \method{} can be layered on top of resolution-aware schedulers---a point we emphasize by applying it to Flux-Dev, which already employs a resolution-dependent schedule.
Compared with downsampling, retraining~\cite{hu2022lora, Cheng2024ResAdapterDC}, and resizer-based inference methods~\cite{elasticdiffusion, scalecrafter}, \method{} offers an efficient and flexible alternative: it requires no parameter updates, does not alter the sampling trajectory, and relies only on a small \update{implementation change} and one-time calibration cost per target resolution (see \Cref{fig:pipeline}).

Our contributions are threefold:
\begin{itemize}
    \item We formalize \emph{conditioning miscalibration} as a distinct source of low-resolution degradation in diffusion/flow-matching models, complementary to architectural, procedural, and schedule-based resolution adaptation.
    \item We introduce \method{}, a training-free conditioning recalibration method that keeps the sampling schedule fixed and re-indexes only the denoiser's noise-level conditioning to restore local forward--reverse consistency. \update{\method{} keeps sampling latency the same and requires only a small implementation change.}
    \item We show that \method{} is lightweight, reusable, and composable with resolution-aware schedulers, yielding consistent improvements for direct low-resolution sampling across several modern diffusion models and datasets.
\end{itemize}

\section{Related Work}
\label{sec:related_work}

\subsubsection{Diffusion Models at Arbitrary Resolutions.}
Most diffusion models are trained at a single, fixed size, but real applications demand flexible outputs. Multiple works have been proposed to adapt pretrained diffusion models to variable resolutions in a training or training-free manner~\cite{freescale, cheap-scaling, foleycrafter, any-size-diffusion, Cheng2024ResAdapterDC, demofusion}. Multidiffusion~\cite{multidiffusion} {\em stitches} multiple overlapping diffusion windows, enabling panoramas and extreme aspect ratios without retraining.
ElasticDiffusion~\cite{elasticdiffusion} separates global structure and local detail during decoding, allowing a pretrained model to scale up or down in resolution while remaining artifact-free. ScaleCrafter~\cite{scalecrafter} enlarges the pretrained diffusion model receptive field on-the-fly via re-dilation and couples it with noise-damped guidance, achieving 4K resolution from a 512p model. These methods highlight that inference-time adaptations can unlock arbitrary-size synthesis, yet none address the \emph{noise-level misalignment} that emerges when the same schedule is reused across resolutions.

\subsubsection{Test-Time Calibration and Denoising Consistency.}

A complementary direction of research focuses on inference-time strategies to calibrate the diffusion process and improve the generation quality.
Chen \emph{et al.}~\cite{importantnoisescheudler} show that default 
linear or cosine
schedules leave a residual signal and propose schedule rescaling to close this train–test gap. WSNR Sampling~\cite{wsnr} refines the schedule to keep a constant weighted SNR across domains, boosting high-res fidelity. 
NAG~\cite{zhong2025mitigating} studies the schedule-vs-actual-noise drift in intermediate states, while we target resolution-induced conditioning miscalibration and performs training-free conditioning remapping.
ScaleCrafter~\cite{scalecrafter} further introduces noise-dependent classifier-free guidance, lowering guidance in early noisy steps and increasing it later. 
High-order ODE solvers such as the DPM-Solver~\cite{dpm, dpmplus} shorten the sampling trajectory while preserving consistency.  
Time-shifting~\cite{timeshifting} reduces exposure bias by reparameterizing the inference-time noise schedule, thereby changing the sampling trajectory to better match the learned denoising dynamics of the model.
Our work is complementary; instead of modifying the global schedule or the sampler, we re-index the timestep embedding to 
{\em fool}
the model into operating at the correct noise level for each resolution.

\subsubsection{Perceptual Effects of Noise Across Resolutions.}
Resolution changes alter how noise corrupts perceptual content.  
Jin \emph{et al.}~\cite{jinentropy} observe that when prompting a pretrained diffusion model to generate images outside their training set, low-res images lose fine details, whereas high-res outputs duplicate objects, and proposed an entropy-based attention scale to mitigate this issue.  
ScaleCrafter~\cite{scalecrafter} and ElasticDiffusion~\cite{elasticdiffusion} report similar artifacts and attribute them to a limited receptive field. Chen \emph{et al.}~\cite{importantnoisescheudler} quantify that a fixed noise percentage degrades 256$\times$256 images far more than 1024
$\times$1024 images, motivating scale-aware schedules, which several subsequent work formalized~\cite{wsnr, timeshifting}.  
These studies underline that identical timesteps correspond to \emph{different} perceptual noise levels across resolutions, a mismatch \method{} explicitly corrects.

\section{Method}
\label{sec:method}

We present \method{}, a training-free test-time calibration method for direct low-resolution sampling.
The key idea is to decouple the \emph{scheduled noise} that defines the sampling trajectory from the \emph{conditioning noise} fed to the denoiser through its timestep/noise embedding.
As illustrated in \Cref{fig:pipeline}, \method{} performs a one-time per-resolution calibration to learn a mapping from scheduled noise (or step index) to a calibrated conditioning noise, which is then cached and reused for unseen prompts at inference.
This design keeps the sampler and its schedule unchanged while adapting the denoiser's conditioning, making \method{} complementary to schedule-based strategies such as time shifting.  We begin by reviewing the flow matching framework (\Cref{sec:preliminaries}).
We then formalize the schedule--conditioning distinction (\Cref{sec:cond_vs_schedule}). Finally, we introduce \method, our training-free method to calibrate the conditioning noise level through coarse-to-fine grid search (\Cref{sec:calibration}). This calibration is performed once per resolution and reused during inference without modifying the model or the noise schedule.

\subsection{Preliminaries: Flow Matching}
\label{sec:preliminaries}
\update{Flow matching~\cite{lipman2022flow} is a training paradigm for generative models that learns a continuous transformation from a simple base distribution $p_0(\mathbf{x})$ (\textit{e.g.}, Gaussian noise) to a complex target distribution $q(\mathbf{x})$. It directly regresses the velocity field of an ordinary differential equation (ODE), enabling simulation-free learning of the generative process. Throughout, we use $\mathbf{x}_0 \sim q$ to denote a clean data sample and $\mathbf{x}_1 \sim p_0$ to denote a noisy sample. The trajectory is defined as a continuous interpolation between $\mathbf{x}_0$ and $\mathbf{x}_1$ along a predefined path.}

The training objective minimizes the discrepancy between a predicted velocity $v_t(\mathbf{x}_t)$ and a target velocity $u_t(\mathbf{x}_t \mid \mathbf{x}_0, \mathbf{x}_1)$, which is analytically derived from the interpolation path:
\begin{equation}
    \mathbb{E}_{t,\, \mathbf{x}_0 \sim q,\, \mathbf{x}_1 \sim p_0}\left[ \left\| v_t(\mathbf{x}_t) - u_t(\mathbf{x}_t \mid \mathbf{x}_0, \mathbf{x}_1) \right\|^2 \right],
\end{equation}
where $\mathbf{x}_t$ follows a time-dependent interpolant between $\mathbf{x}_0$ and $\mathbf{x}_1$, such as:
\begin{equation}
\begin{aligned}
    \mathbf{x}_t &= (1 - t)\,\mathbf{x}_0 + t\,\mathbf{x}_1, \\
    u_t(\mathbf{x}_t \mid \mathbf{x}_0,\mathbf{x}_1) &= \frac{d\mathbf{x}_t}{dt} = \mathbf{x}_1 - \mathbf{x}_0, \\
    \mathbf{x}_t \mid \mathbf{x}_0 &\sim \mathcal{N}\!\big((1-t)\,\mathbf{x}_0,\, t^2 \mathbf{I}\big).
\end{aligned}
\end{equation}

This framework has been adopted in recent diffusion transformers such as Stable Diffusion 3~\cite{Esser2024ScalingRF} and Flux~\cite{flux2024}, which we study in this paper. These models generate images by integrating the learned velocity field over a predefined noise schedule $\{\sigma_t\}$ (starting from $\mathbf{x}_1 \sim p_0$ and integrating toward $\mathbf{x}_0$), where our method intervenes at test time to correct resolution-specific inconsistencies.

\begin{algorithm}[t]
\caption{Coarse-to-Fine Conditioning Noise Calibration 
}
\label{alg:coarse_to_fine}
\begin{algorithmic}[1]
\REQUIRE Model $\phi$, clean image $\mathbf{x}_0$, noise schedule $\{\sigma_t\}$, timestep $t$, upper bound $\hat{\sigma}_{t+1}^*$
\STATE $\mathbf{x}_{t+1} \gets \text{AddNoise}(\mathbf{x}_0, \sigma_{t+1})$
\STATE $\mathbf{x}_t \gets \text{AddNoise}(\mathbf{x}_0, \sigma_t)$
\STATE Initialize $\hat{\sigma}_t^* \gets \sigma_t$, \quad $\mathcal{L}^* \gets \|\hat{\mathbf{x}}_t - \mathbf{x}_t\|^2$
\STATE Set coarse search range: 
$
\tilde{\sigma} \in [\max(0, \sigma_t - \epsilon_c), \min(\hat{\sigma}_{t+1}^*, \sigma_t + \epsilon_c)]
$
\FORALL{$\tilde{\sigma} \in$ coarse candidates (stride $\delta_c$)}
    \STATE $\hat{\mathbf{x}}_t \gets \mathbf{x}_{t+1} + \phi(\mathbf{x}_{t+1}, \tilde{\sigma}) \cdot \Delta t$
    \STATE $\mathcal{L} \gets \|\hat{\mathbf{x}}_t - \mathbf{x}_t\|^2$
    \IF{$\mathcal{L} < \mathcal{L}^*$}
        \STATE $\hat{\sigma}_t^* \gets \tilde{\sigma}$, \quad $\mathcal{L}^* \gets \mathcal{L}$
    \ENDIF
\ENDFOR
\STATE Set fine search range: 
$
\tilde{\sigma} \in [\max(0, \hat{\sigma}_t^* - \epsilon_f), \min(\hat{\sigma}_{t+1}^*, \hat{\sigma}_t^* + \epsilon_f)]
$
\FORALL{$\tilde{\sigma} \in$ fine candidates (stride $\delta_f$, $\delta_f$ $<$ $\delta_c$)}
    \STATE Repeat lines 6–10
\ENDFOR
\RETURN Optimized $\hat{\sigma}_t^*$
\end{algorithmic}
\end{algorithm}

\subsection{Resolution-Aware Test-Time Calibration}
\label{sec:cond_vs_schedule}

To clarify the intervention of \method{}, we distinguish two roles played by the noise level during inference: (i) the \emph{scheduled noise level}, \update{denoted as $\sigma_t$}  which determines the numerical sampling trajectory, and (ii) the \emph{conditioning noise level}, \update{denoted as $\hat{\sigma}_t$}, which is used only as the model's timestep/noise input.
This distinction is central to \method{}: unlike schedule-based methods, we do not modify the sampling trajectory itself (see \Cref{fig:pipeline}).

\update{During training, $\hat{\sigma}_t$ is set to be equal to $\sigma_t$ to inform the model of the input corruption level. However, we point out that at inference on unseen lower resolutions, the same scheduled noise level $\sigma_t$ induces stronger effective corruption (see \Cref{fig:pipeline}), leading to a train--test mismatch and degraded generation quality. To address this, prior work~\cite{timeshifting} proposed to alter the sampling noise scheduler by remapping the current sample at time $t$ ($\mathbf{x}_{t}$) to a more appropriate timestep ($\mathbf{x}_{\tilde{t}}$). In this work, we propose instead to recalibrate the conditioning noise level to better reflect the effective corruption. Concretely, we introduce a resolution-dependent mapping $g_r$ that maps the scheduled noise level to a recalibrated conditioning level, which we use to adjust the noise-level conditioning during inference.}
\update{We rewrite the Euler sampler update rule as:}

\begin{equation}
    \mathbf{x}_t
    =
    \mathbf{x}_{t+1}
    +
    \phi(\mathbf{x}_{t+1}, \hat{\sigma}_t)\,(\sigma_{t+1} - \sigma_t).
    \label{eq:conditioned_sampling}
\end{equation}
where $\phi$ denotes the pretrained flow-matching model and $\{\sigma_t\}_{t=0}^{T}$ is the scheduled noise sequence used at inference. \update{For simplicity, we remove the text conditioning from the equation.}
Equation~\eqref{eq:conditioned_sampling} makes the separation clear: the scheduled noise $\sigma_t$ still defines the solver step, while the conditioning noise $\hat{\sigma}_t$ controls what noise level the network is instructed to denoise.
As a result, \method{} keeps the scheduled sequence $\{\sigma_t\}_{t=0}^{T}$ fixed and calibrates only the conditioning sequence $\{\hat{\sigma}_t\}_{t=0}^{T-1}$.
Formally, for each target resolution $r$, we optimize a lightweight mapping
\begin{equation}
    \hat{\sigma}_t^{*} = g_r(\sigma_t),
    \label{eq:noise_shift_mapping}
\end{equation}
and then sample using \eqref{eq:conditioned_sampling} with $\hat{\sigma}_t = \hat{\sigma}_t^{*}$.

\update{This yields a minimal, training-free test-time adapter that improves generation quality at unseen lower resolutions without introducing any inference overhead. Since \method{} focuses on recalibrating the noise-conditioning level, it is complementary} to schedule-based approaches and can therefore be composed with them (e.g., on top of the resolution-aware scheduler in Flux-Dev).

\subsection{Coarse-to-Fine Search for Calibrated Conditioning Levels}
\label{sec:calibration}

At each reverse step, the model receives not only the current noisy latent $\mathbf{x}_{t+1}$ but also a noise-level (or timestep) input that indicates the expected corruption level.
This conditioning allows the model to adapt its velocity prediction to the amount of noise present in the current state.
In our formulation, the conditioning noise $\hat{\sigma}_t$ is passed directly as this model input.
Therefore, by modifying $\hat{\sigma}_t$ at test time, we can explicitly control how the model interprets the noise level in $\mathbf{x}_{t+1}$, enabling resolution-aware correction without changing the model weights or the scheduled sampling trajectory.

\paragraph{Problem Setup.}
Let $\phi$ denote a trained flow-matching diffusion model with a predefined noise schedule ${\sigma_t}{t=0}^{T}$. At each timestep $t$, the model receives a noisy latent $\mathbf{x}_{t+1}$ and a noise-level embedding derived from a conditioning noise $\hat{\sigma}_t$. This embedding informs the network how much noise to expect and guides its prediction of the velocity field $\phi(\mathbf{x}_{t+1}, \hat{\sigma}_t)$.

In standard sampling, $\hat{\sigma}_t = \sigma_t$ follows the original noise schedule. In our approach, we allow $\hat{\sigma}_t$ to differ from $\sigma_t$ to correct forward–reverse misalignment at test time. By tuning $\hat{\sigma}_t$, we can adapt the model’s denoising behavior to different resolutions without changing the learned weights or sampling procedure.

Given a noisy latent input $\mathbf{x}_{t+1}$, we expect $\mathbf{x}_t$ to match the forward sample obtained by applying Gaussian noise to a clean latent $\mathbf{x}_0$. Our goal is to find a conditioning noise level $\hat{\sigma}_t$ such that the denoised prediction $\hat{\mathbf{x}}_t$ minimizes the discrepancy from $\mathbf{x}_t$:
\begin{equation}
    \hat{\mathbf{x}}_t = \mathbf{x}_{t+1} + \phi(\mathbf{x}_{t+1}, \hat{\sigma}_t) \cdot \Delta t, \quad \text{where } \Delta t = \sigma_{t+1} - \sigma_t,
\end{equation}
\begin{equation}
    \hat{\sigma}_t^* = \arg\min_{\hat{\sigma}} \; \| \hat{\mathbf{x}}_t - \mathbf{x}_t \|^2.
\end{equation}

\begin{algorithm}[t]
\caption{Euler Sampling with Calibrated Conditioning}
\label{alg:euler_new_sigmas}
\begin{algorithmic}[1]
    \REQUIRE Model $\phi$, initial latent $\mathbf{x}_T$, noise schedule $\{\sigma_t\}_{t=0}^{T}$, calibrated conditioning $\{\hat{\sigma}_t^*\}_{t=0}^{T-1}$
    \FOR{$t = T-1$ down to $0$}
        \STATE $\mathbf{v}_t \gets \phi(\mathbf{x}_{t+1}, \hat{\sigma}_t^*)$
        \STATE $\mathbf{x}_t \gets \mathbf{x}_{t+1} + \mathbf{v}_t\,(\sigma_{t+1} - \sigma_t)$
    \ENDFOR
    \RETURN $\mathbf{x}_0$
\end{algorithmic}
\end{algorithm}

\paragraph{Coarse-to-Fine Search Strategy.}
We estimate $\hat{\sigma}_t^*$ via a coarse-to-fine grid search that minimizes the one-step reverse error at each timestep. Starting from the final step $t = T{-}1$, we proceed backward in time. For each $t$, we first evaluate the default conditioning $\hat{\sigma}_t = \sigma_t$, then perform a coarse sweep with a larger stride $\delta_c$ to identify a promising region. A subsequent fine-grained search with a smaller stride $\delta_f$ refines the estimate within a narrow window.

To ensure monotonic consistency with the diffusion trajectory, we constrain all candidate values to lie in $[0, \hat{\sigma}_{t+1}^*]$. This backward recursion allows us to progressively calibrate the denoising inputs while respecting the ordering of noise levels. The full procedure is detailed in Algorithm~\ref{alg:coarse_to_fine}.

We perform this calibration over a small set of
image-text pairs for each resolution. The resulting schedule $\{\hat{\sigma}_t^*\}$ is resolution-specific and cached for reuse at inference.

\paragraph{Inference with Calibrated Conditioning.}
At inference time, we sample using the original noise schedule $\{\sigma_t\}$ to preserve the intended diffusion trajectory. However, at each timestep $t$, we replace the model input conditioning with the precomputed value $\hat{\sigma}_t^*$ obtained from the calibration procedure. These values are resolution-specific but fixed across inputs, and require no model retraining or architectural changes. As shown in Algorithm~\ref{alg:euler_new_sigmas}, the update step becomes:
\begin{equation}
    \mathbf{x}_t = \mathbf{x}_{t+1} + \phi(\mathbf{x}_{t+1}, \hat{\sigma}_t^*) \cdot (\sigma_{t+1} - \sigma_t).
\end{equation}

\section{Experimental Setup}
\label{sec:exp_setup}

\subsubsection{Implementation Details.} For small-scale lightweight calibration, we randomly sample 200 web-crawled images from SBU~\cite{Ordonez2011Im2TextDI}. These samples are only used to estimate resolution-specific conditioning noise levels. For all experiments, we use a coarse-to-fine search strategy with a coarse search window $\epsilon_c = 0.1$ and a fine search window $\epsilon_f = 0.01$. 
Experiments are conducted on NVIDIA A40s. On a single node with 8 A40 GPUs, calibrating on 200 128$\times$128 images takes around 13 minutes for SD3, 25 minutes for SD3.5.

\vspace{-0.1in}
\subsubsection{Datasets.}
We evaluate on LAION-COCO~\cite{LAIONcoco2024}, a diverse subset of LAION-5B~\cite{Schuhmann2022LAION5BAO} containing multi-resolution images with BLIP-generated captions~\cite{Li2022BLIPBL}. It includes a broad spectrum of content, such as objects, people, and natural scenes. We also include an evaluation on the CelebA benchmark~\cite{liu2015faceattributes}, a face-centric image–text dataset.

\vspace{-0.1in}
\subsubsection{Metrics.}
We evaluate models with standard text-to-image generation metrics: CLIP Score~\cite{Hessel2021CLIPScoreAR} for text–image alignment and Frechet Inception Distance (FID)~\cite{Heusel2017GANsTB} for the distance between feature distributions of real and generated images. Following prior works~\cite{elasticdiffusion, scalecrafter, Cheng2024ResAdapterDC}, we sample 10,000 images for each of the benchmarks. While we disable CFG (\textit{i.e.}, set the guidance scale to 1.0) during calibration, we set 5.0 for SD3, 4.5 for SD3.5, and 3.0 for Flux-Dev during inference.

\vspace{-0.1in}
\subsubsection{Models.}
We validate our method on flow-matching diffusion models with various configurations. We integrate it to Stable Diffusion 3~\cite{Esser2024ScalingRF} with a linear noise schedule and Stable Diffusion 3.5~\cite{Esser2024ScalingRF} with a non-linear schedule that spends more time in low- or high-noise regimes. 
We also experiment with Flux-Dev~\cite{flux2024}, an open-sourced distilled flow-matching DiT model that incorporates a resolution-aware scheduler that applies a shift parameter based on image size.

\section{Experimental Results}
\label{sec:exp_results}
This section evaluates \method{} with fixed linear and non-linear noise sampling schedulers in Sec.~\ref{sec:exp_sd3_sd35}, analyzes calibrated noise conditioning in Sec.~\ref{sec:exp_calibrated_noise}, validates the effectiveness on top of time-shifting noise scheduler in Sec.~\ref{sec:exp_time_shifting}, ablate the impact of the size of calibration set in Sec.~\ref{sec:exp_ablation}, and provide qualitative examples in Sec.~\ref{sec:exp_qualitative}.

\subsection{Fixed Linear or Non-linear Noise Schedules}
\label{sec:exp_sd3_sd35}
We evaluate \method{} using two flow-matching diffusion models with fixed noise schedules (See Figure~\ref{alg:euler_new_sigmas}). Stable Diffusion 3 (SD3) uses a linear noise schedule shared across resolutions, while Stable Diffusion 3.5 (SD3.5) adopts a fixed non-linear schedule that biases sampling toward low- or high-noise regimes.

We conduct inference with calibrated noise schedulers across resolutions with SD3 and SD3.5. As shown in Table~\ref{tab:merged_main_results}, \method{} consistently improves CLIP Score and FID for both SD3 and SD3.5 across non-default resolutions ranging from 128$\times$128 to 768$\times$768. 
\method{} provides measurable quality gains through simple post-hoc calibration when applied to lower resolutions.

\begin{table*}[t]
\centering
\caption{\textbf{Quantitative evaluation across datasets and resolutions.}
We report CLIP score (↑) and FID (↓) for SD3 and SD3.5 with and without \method{} on CelebA and LAION-COCO. }
\label{tab:merged_main_results}
\small
\setlength{\tabcolsep}{4pt}
\scalebox{0.82}{
\begin{tabular}{l l l c c c c c}
\toprule
 &  & & \multicolumn{2}{c}{\textbf{SD3}} & \multicolumn{2}{c}{\textbf{SD3.5}} \\
 \cmidrule(lr){4-5}\cmidrule(lr){6-7}
\textbf{Dataset} & \textbf{Res.} & & \textbf{CLIP$\uparrow$} & \textbf{FID$\downarrow$} & \textbf{CLIP$\uparrow$} & \textbf{FID$\downarrow$}  \\
\midrule
\multirow{8}{*}{CelebA}
& \multirow{2}{*}{128$\times$128}
& Base   & 21.07 & 320.45 & 19.01 & 386.86 \\
&  & +Ours  & \textbf{21.86}$^{\color{violet}+3.75\%}$ & \textbf{311.89}$^{\color{violet}-2.67\%}$
           & \textbf{20.56}$^{\color{violet}+8.15\%}$ & \textbf{374.74}$^{\color{violet}-3.13\%}$ &  \\
\cmidrule(lr){2-8}
& \multirow{2}{*}{256$\times$256}
& Base   & 22.14 & 291.26 & 19.96 & 359.00  \\
&  & +Ours  & \textbf{23.76}$^{\color{violet}+7.32\%}$ & \textbf{252.61}$^{\color{violet}-13.27\%}$
           & \textbf{20.21}$^{\color{violet}+1.25\%}$ & \textbf{271.52}$^{\color{violet}-24.37\%}$ &  \\
\cmidrule(lr){2-8}
& \multirow{2}{*}{512$\times$512}
& Base   & 25.54 & 128.62 & 22.27 & 292.42 \\
&  & +Ours  & \textbf{25.74}$^{\color{violet}+0.78\%}$ & \textbf{123.14}$^{\color{violet}-4.26\%}$
           & \textbf{23.51}$^{\color{violet}+5.57\%}$ & \textbf{270.36}$^{\color{violet}-7.54\%}$ &  \\
\cmidrule(lr){2-8}
& \multirow{2}{*}{768$\times$768}
& Base   & 27.02 & 93.66 & 26.68 & 135.84 \\
&  & +Ours  & \textbf{27.03}$^{\color{violet}+0.04\%}$ & \textbf{93.14}$^{\color{violet}-0.56\%}$
           & \textbf{26.91}$^{\color{violet}+0.86\%}$ & \textbf{127.17}$^{\color{violet}-6.38\%}$ &  \\
\midrule
\multirow{8}{*}{LAION-COCO}
& \multirow{2}{*}{128$\times$128}
& Base   & 19.80 & 203.23 & 19.18 & 310.40 \\
&  & +Ours  & \textbf{21.07}$^{\color{violet}+6.41\%}$ & \textbf{170.93}$^{\color{violet}-15.89\%}$
           & \textbf{19.75}$^{\color{violet}+2.97\%}$ & \textbf{276.90}$^{\color{violet}-10.79\%}$ &  \\
\cmidrule(lr){2-8}
& \multirow{2}{*}{256$\times$256}
& Base   & 22.24 & 159.13 & 19.46 & 256.31 \\
&  & +Ours  & \textbf{23.28}$^{\color{violet}+4.68\%}$ & \textbf{130.84}$^{\color{violet}-17.78\%}$
           & \textbf{20.23}$^{\color{violet}+3.96\%}$ & \textbf{175.14}$^{\color{violet}-31.67\%}$ &  \\
\cmidrule(lr){2-8}
& \multirow{2}{*}{512$\times$512}
& Base   & 28.52 & 76.49 & 22.26 & 203.55 \\
&  & +Ours  & \textbf{28.61}$^{\color{violet}+0.32\%}$ & \textbf{75.86}$^{\color{violet}-0.82\%}$
           & \textbf{23.41}$^{\color{violet}+5.17\%}$ & \textbf{174.20}$^{\color{violet}-14.42\%}$ &  \\
\cmidrule(lr){2-8}
& \multirow{2}{*}{768$\times$768}
& Base   & \textbf{30.10} & 55.13 & 31.15 & 45.05 \\
&  & +Ours  & \textbf{30.10}$^{\color{gray}+0.00\%}$ & \textbf{55.07}$^{\color{violet}-0.11\%}$
           & \textbf{31.28}$^{\color{violet}+0.42\%}$ & \textbf{42.05}$^{\color{violet}-6.66\%}$ &  \\
\bottomrule
\end{tabular}
}
\end{table*}

\begin{figure*}[t]
    \centering
    \includegraphics[width=0.8\linewidth]{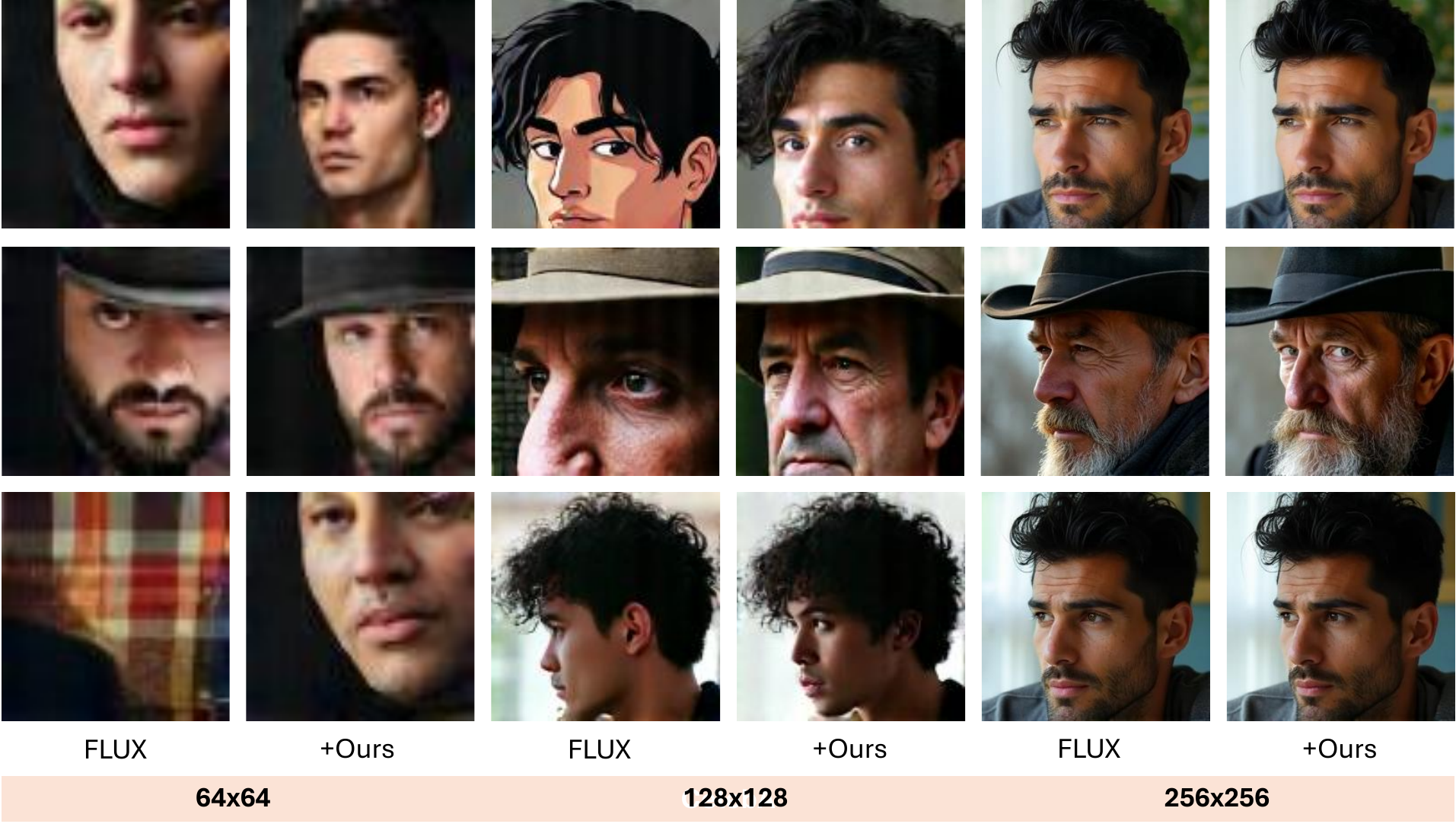}
    \includegraphics[width=0.8\linewidth]{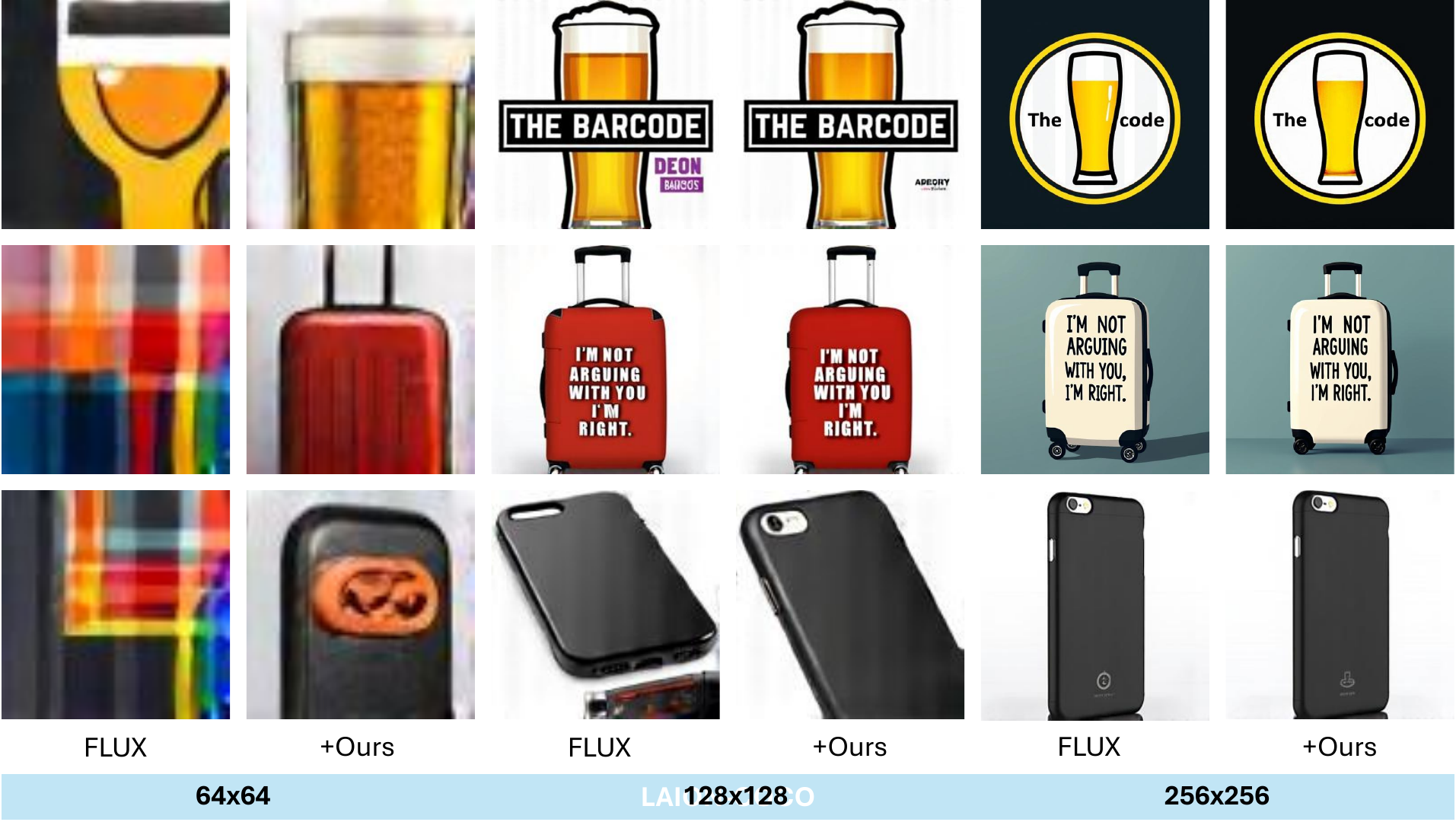}
    \caption{\textbf{Qualitative comparison of Flux-Dev.} Generated image examples before and after applying \method{} are on CelebA (up) and LAION-COCO (down).}
    \label{fig:qualitative_flux}
\end{figure*}

\begin{figure}[h]
    \centering
    \includegraphics[width=0.48\linewidth]{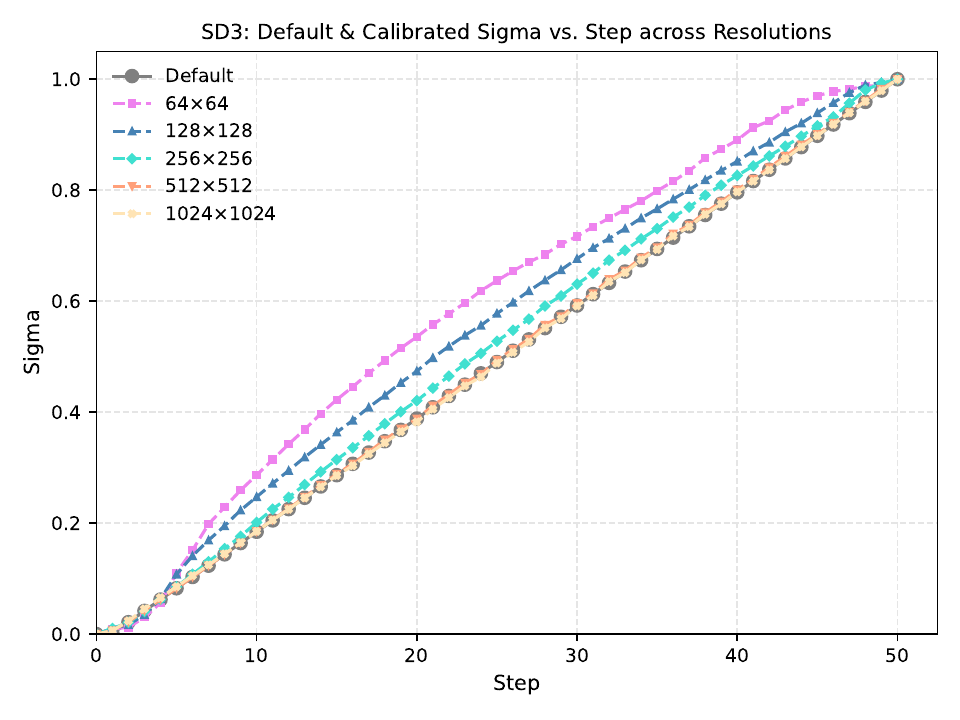}
    \includegraphics[width=0.48\linewidth]{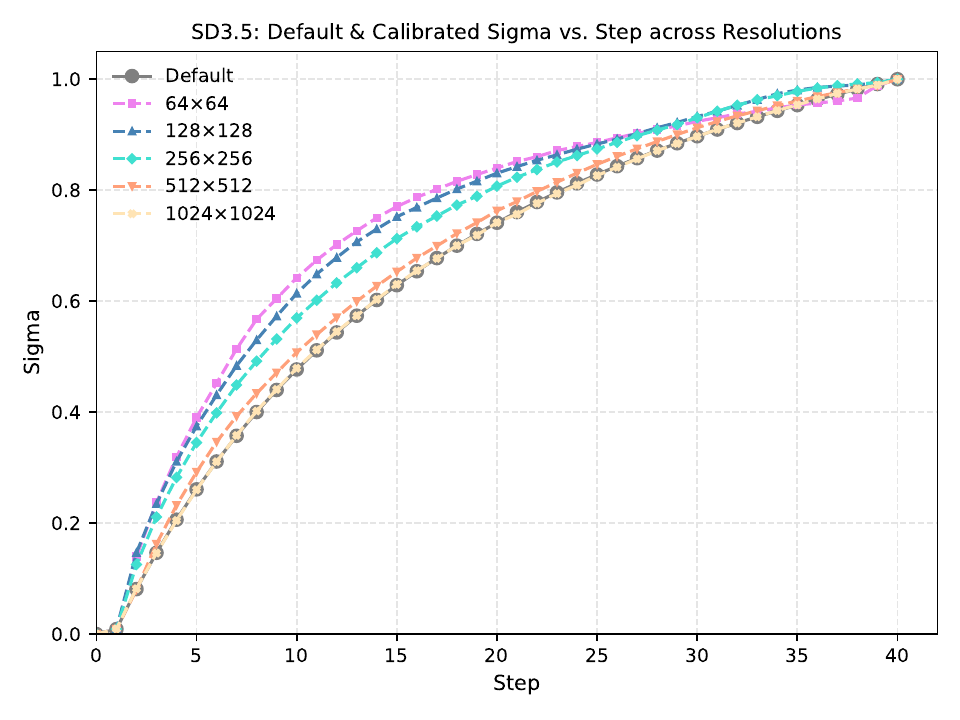}
    \caption{\textbf{Calibrated conditioning noise levels across resolutions.} 
We plot the default sampling noise schedule (gray) alongside the resolution-specific calibrated conditioning $\hat{\sigma}_t$ for SD3 (left) and SD3.5 (right). At the default resolution (1024$\times$1024), the curves align closely. At lower resolutions, the optimal $\hat{\sigma}_t$ curves consistently deviate upward, reflecting a need for stronger conditioning to compensate for perceptual degradation. 
}

\label{fig:sd-new-sigma}
\end{figure}

\subsection{Calibrated Noise Conditioning}
\label{sec:exp_calibrated_noise}
Figure~\ref{fig:sd-new-sigma} visualizes the calibrated conditioning noise levels $\hat{\sigma}_t$ from our method across resolutions for SD3 (left) and SD3.5 (right). At the default resolution of 1024$\times$1024, the calibrated curves closely match the original schedule, confirming that the forward–reverse alignment is intact at the training resolution. In contrast, for lower resolutions, optimal $\hat{\sigma}_t$ trajectories consistently shift above the default schedule, indicating that higher conditioning noise levels are needed to correct resolution-specific misalignment.

This behavior supports our core hypothesis: the same sampling noise level $\sigma_t$ has resolution-dependent perceptual effects, which can be effectively mitigated by adjusting only the conditioning noise level during inference. Notably, the magnitude of deviation from the default schedule increases as resolution decreases, aligning with the observed degradation in visual quality. This trend holds across both SD3 and SD3.5 despite their differing noise schedule shapes.

\begin{table}[t]
\centering
\caption{\textbf{NoiseShift is complementary to time-shifting.}Flux-Dev (Base) incorporates a resolution-aware time-shift scheduler and 
serves as a strong schedule-based baseline; +Ours shows the additional gain 
from NoiseShift's conditioning recalibration on top of this scheduler. We 
report CLIP score (↑) and FID (↓) on LAION-COCO and CelebA. Superscript 
percentages show relative improvements (violet) or degradations (gray).}
\label{tab:flux_laion_celeb_main}
\small
\setlength{\tabcolsep}{2pt}
\scalebox{0.8}{
\begin{tabular}{l l c c c c}
\toprule
 & & \multicolumn{2}{c}{\textbf{LAION-COCO}} & \multicolumn{2}{c}{\textbf{CelebA}} \\
\cmidrule(lr){3-4}\cmidrule(lr){5-6}
 & & CLIP↑ & FID↓ & CLIP↑ & FID↓ \\
\midrule
\multirow{2}{*}{64$\times$64}
 & Base   & 25.58  & 119.60  
 & 27.70  & 182.96  \\
 & +Ours  & \textbf{25.81}$^{\color{violet}+0.90\%}$ & \textbf{113.32}$^{\color{violet}-4.80\%}$
          & \textbf{27.77}$^{\color{violet}+0.25\%}$ & \textbf{177.03}$^{\color{violet}-4.22\%}$ \\
\midrule
\multirow{2}{*}{128$\times$128}
 & Base   & 30.74  & 48.00  
 & 28.75  & 90.62  \\
 & +Ours  & \textbf{30.83}$^{\color{violet}+0.29\%}$ & \textbf{47.45}$^{\color{violet}-1.15\%}$
          & \textbf{28.64}$^{\color{gray}-0.38\%}$ & \textbf{87.63}$^{\color{violet}-3.30\%}$ \\
\midrule
\multirow{2}{*}{256$\times$256}
 & Base   & 32.30  & 26.72  
 & \textbf{27.89}  & 56.33  \\
 & +Ours  & \textbf{32.33}$^{\color{violet}+0.09\%}$ & \textbf{25.82}$^{\color{violet}-3.37\%}$
          & \textbf{27.89}$^{\color{gray}+0.00\%}$ & \textbf{55.52}$^{\color{violet}-1.44\%}$ \\
\midrule
\multirow{2}{*}{512$\times$512}
 & Base   & \textbf{32.55}  & 20.13  
 & 28.44  & 87.86  \\
 & +Ours  & \textbf{32.55}$^{\color{gray}+0.00\%}$ & \textbf{19.62}$^{\color{violet}-2.53\%}$
          & \textbf{28.45}$^{\color{violet}+0.04\%}$ & \textbf{86.45}$^{\color{violet}-1.60\%}$ \\
\midrule
\multirow{2}{*}{768$\times$768}
 & Base   & 32.58 & 19.10  
 & 27.90  & 75.52  \\
 & +Ours  & \textbf{32.60}$^{\color{violet}+0.06\%}$ & \textbf{19.03}$^{\color{violet}-0.37\%}$
          & \textbf{28.10}$^{\color{violet}+0.72\%}$ & \textbf{72.10}$^{\color{violet}-4.53\%}$ \\
\bottomrule
\end{tabular}
}
\end{table}

\begin{figure}[t]
    \centering
    \includegraphics[width=0.48\linewidth]{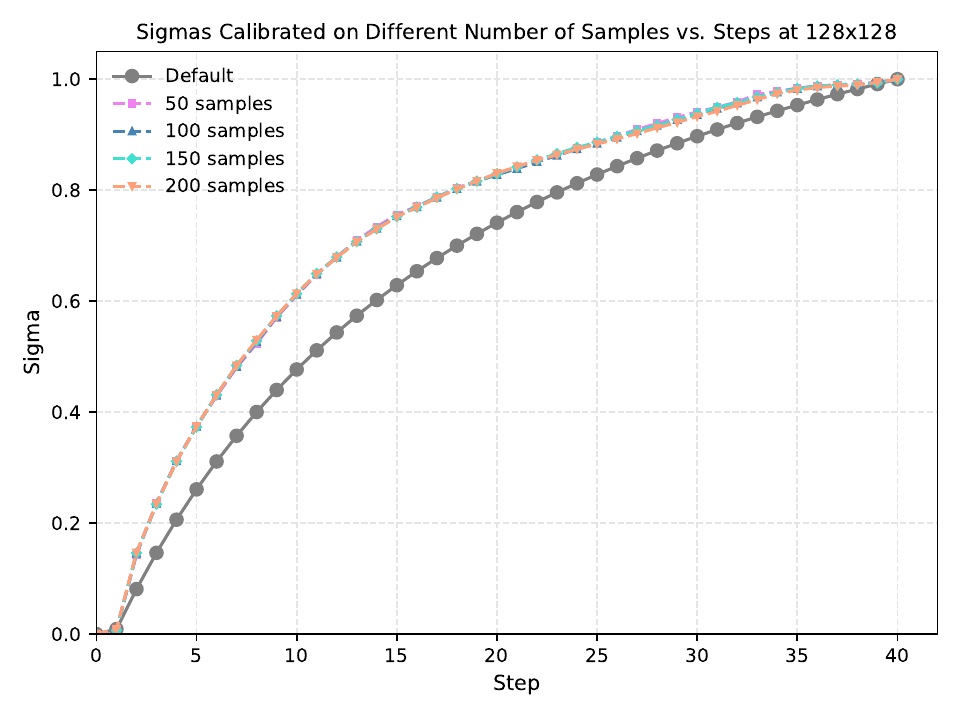}
    \includegraphics[width=0.48\linewidth]{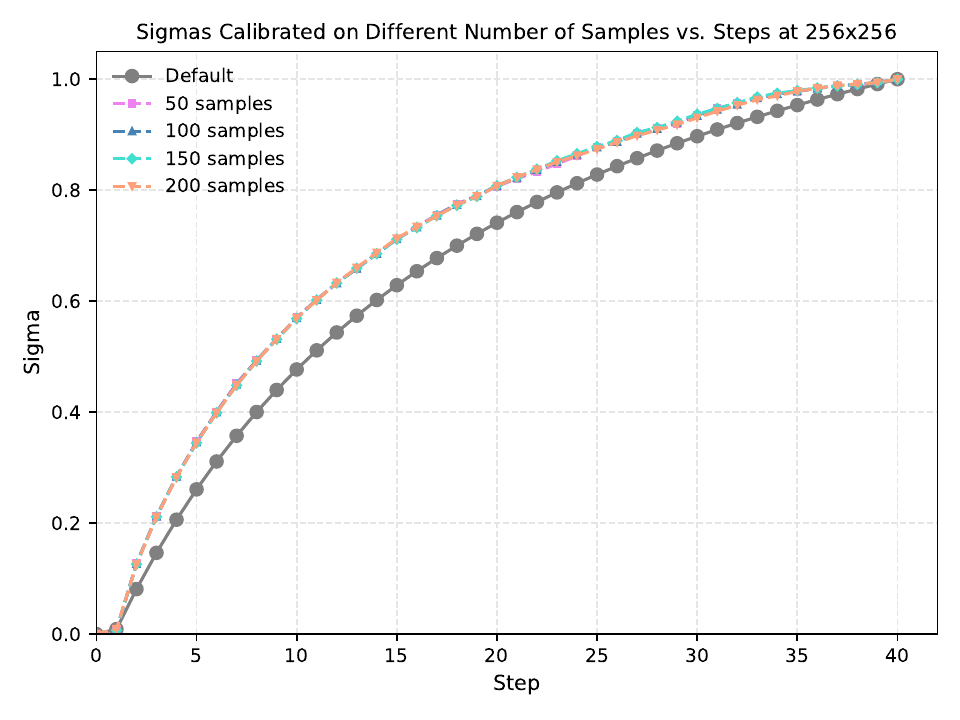}
    \caption{\textbf{Ablation studies.} Ablation studies on the number of samples used during calibration and the new sigmas obtained at 128$\times$128 and 256$\times$256.}
    \label{fig:ablation_num_samples}
\end{figure}

\subsection{Combining NoiseShift with Time-Shifting Schedulers}
\label{sec:exp_time_shifting}
We further evaluate \method{} on Flux-Dev~\cite{flux2024}, a distilled flow-matching DiT model that incorporates a resolution-aware time-shift parameter. 
The noise schedule in Flux-Dev shifts depending on the target resolution: higher-resolution images retain larger $\sigma_t$ values for longer, effectively extending their time in high-signal regimes.

While Flux-Dev adjusts the actual sampling schedule based on resolution, \method{} instead calibrates the \emph{conditioning input} to the denoiser without altering the forward noise schedule itself. In that sense, our approach operates as a lightweight test-time correction, adjusting the model’s expectations without modifying its architecture or training.

As shown in Table~\ref{tab:flux_laion_celeb_main}, \method{} improves the quality of images generated by Flux-Dev from 64$\times$64 to 768$\times$768 on LAION-COCO and CelebA. Although gains are modest, they are notable given that Flux-Dev is a distilled model with a resolution-aware noise sampler, and our method requires very low cost.

\subsection{Ablation Studies}
\label{sec:exp_ablation}
We conduct an ablation study on the number of samples used in the 
\method{} calibration. As shown in Figure~\ref{fig:ablation_num_samples}, even though the number of samples varies from 50 to 200, the calibrated sigmas always converge in almost the same range. Given the fact that the calibration already has a very low computation cost, reducing the number of samples can lead to an almost free re-calibration, which improves the low-resolution image generation quality.

\subsection{Qualitative Results}
\label{sec:exp_qualitative}
\begin{figure}[t]
    \centering
    \includegraphics[width=0.45\linewidth]{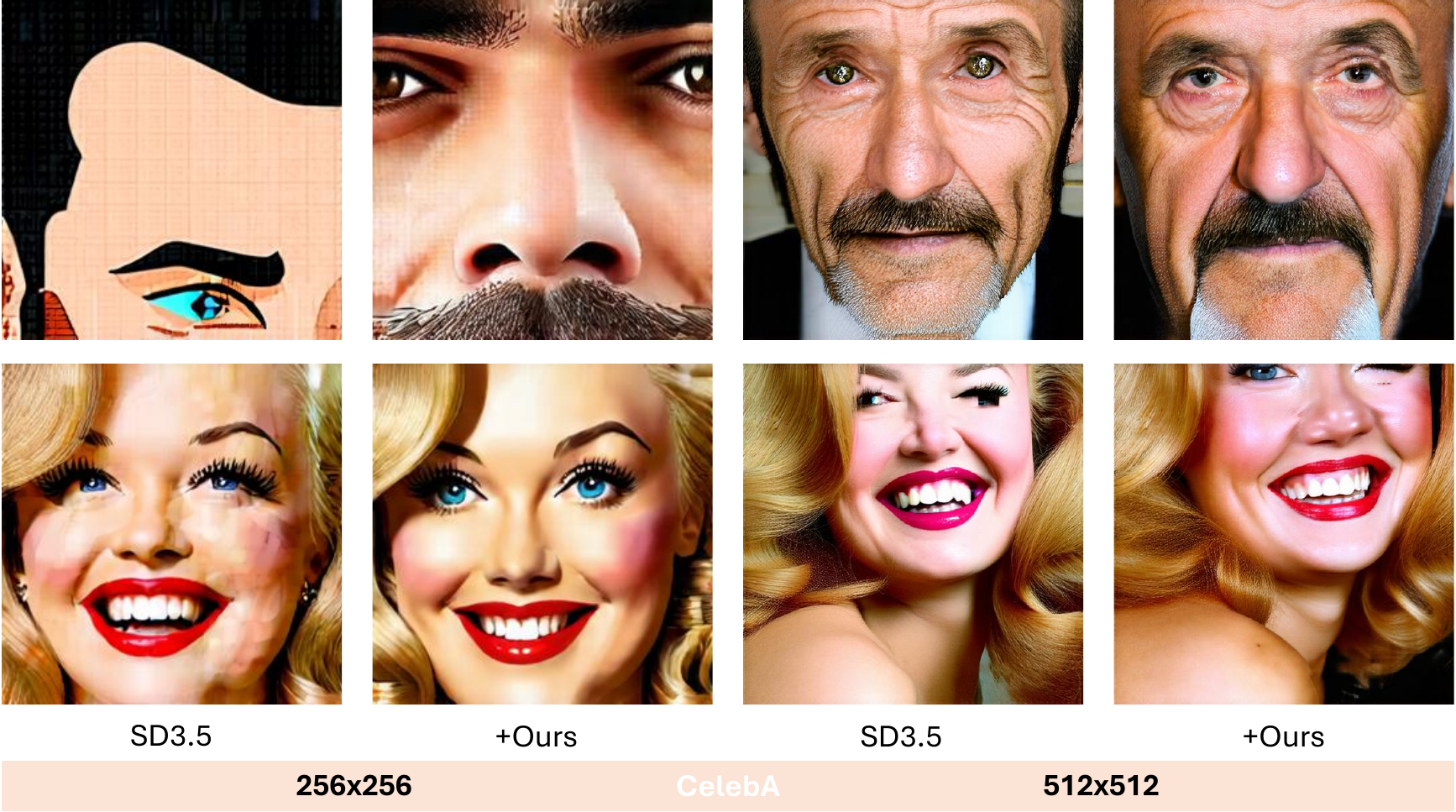}
    \includegraphics[width=0.45\linewidth]{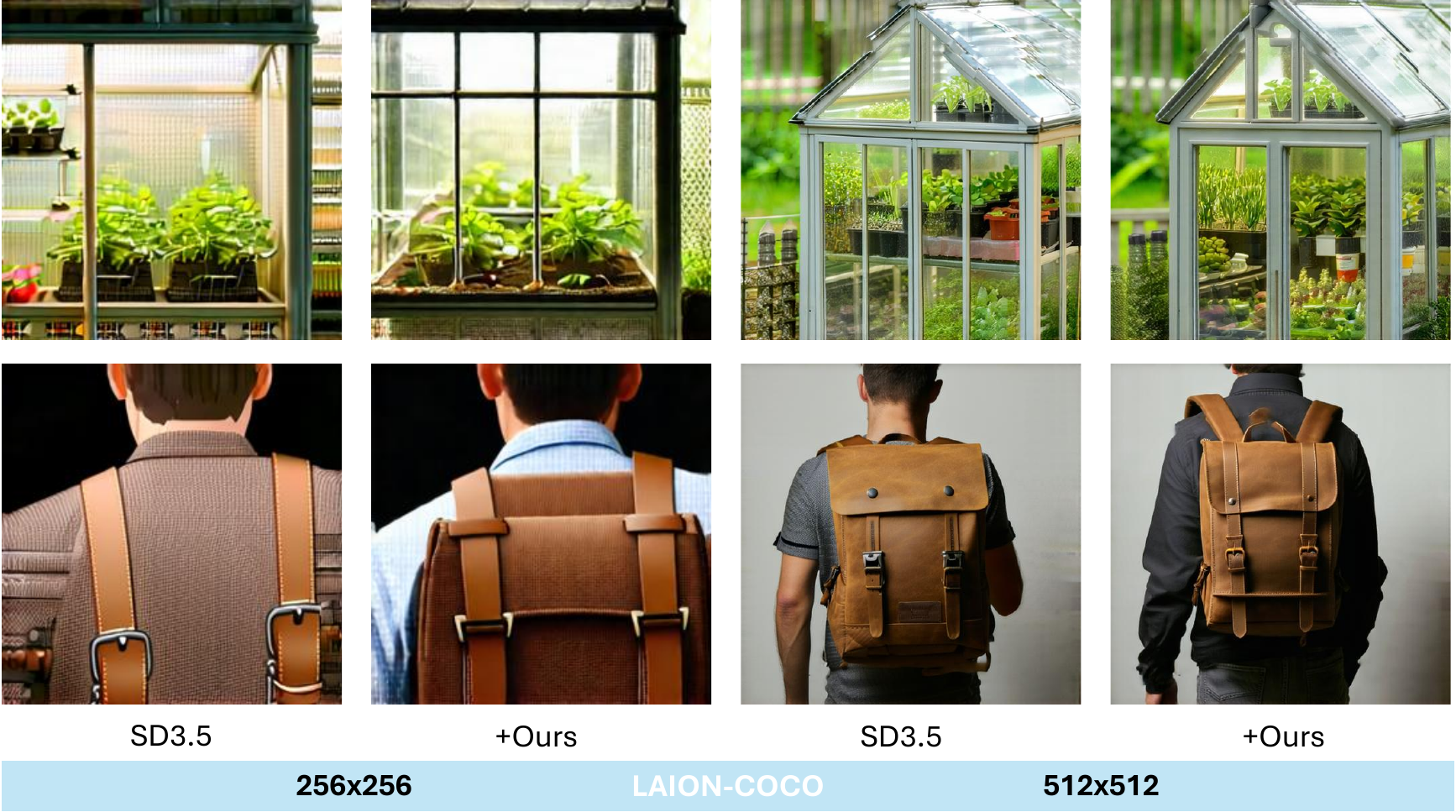}
    \caption{\textbf{Qualitative comparison of SD3.5}. Generated image examples before and after applying \method{} are on CelebA (left) and LAION-COCO (right).}
    \label{fig:qualitative_sd35}
\end{figure}

\begin{figure}[t]
    \centering
    \includegraphics[width=0.45\linewidth]{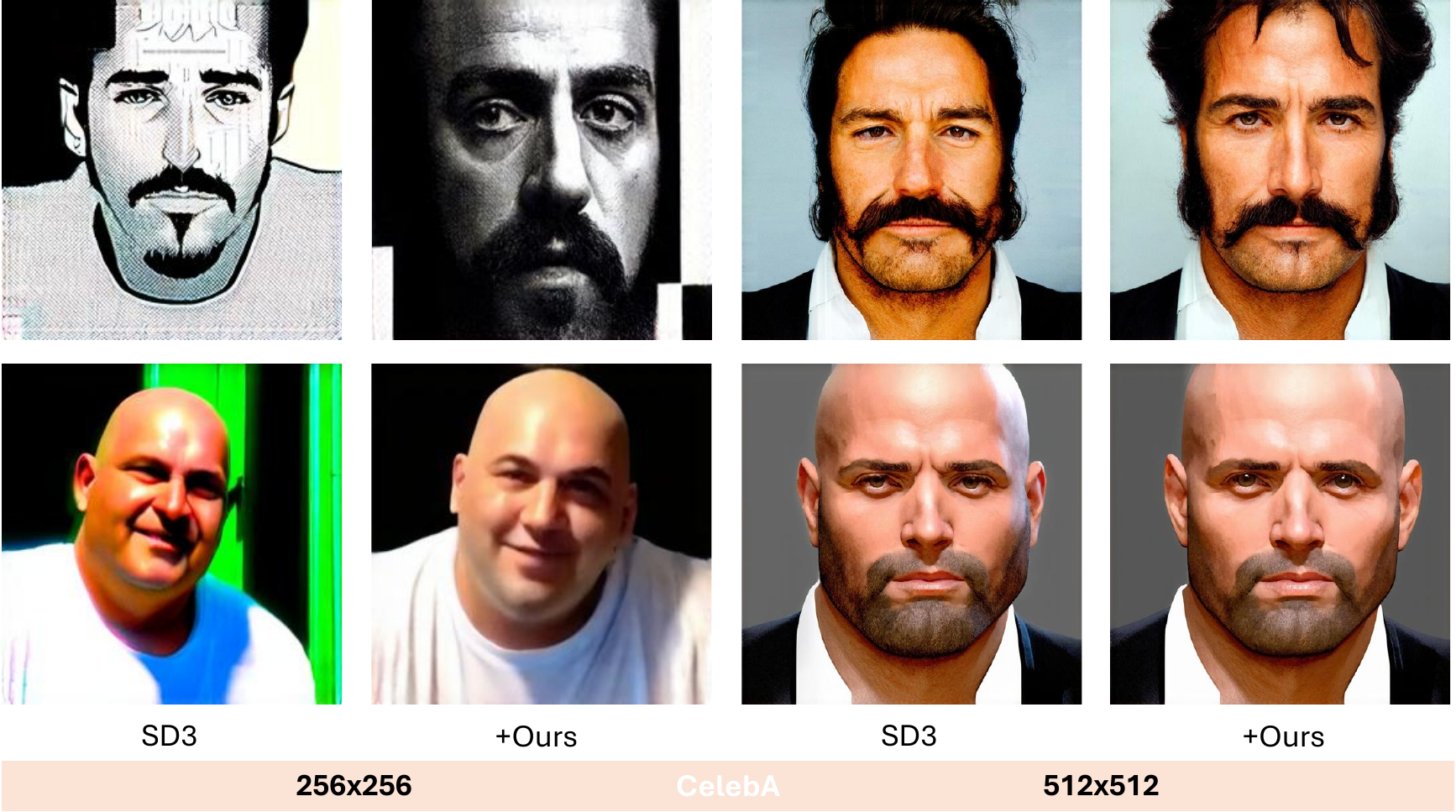}
    \includegraphics[width=0.45\linewidth]{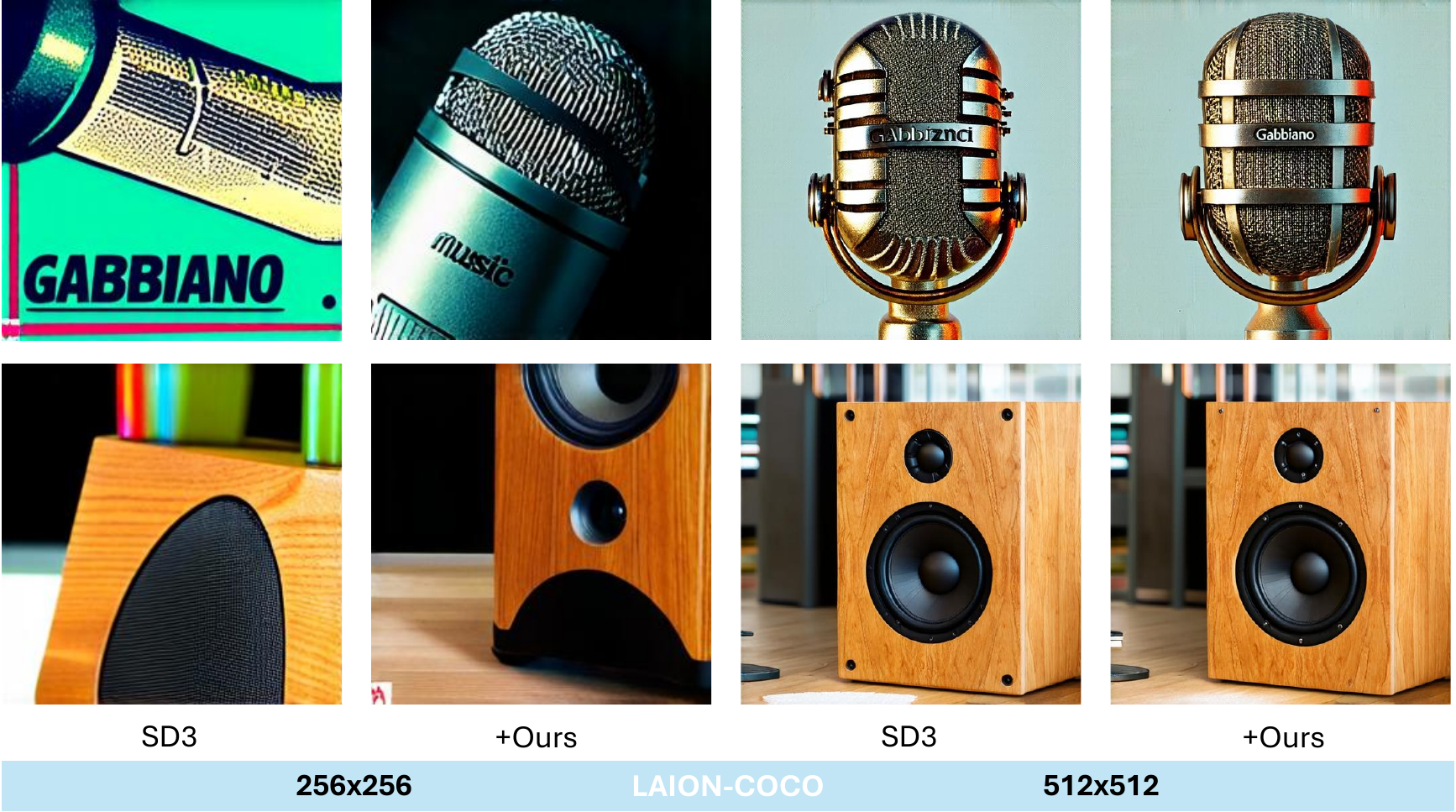}
    \caption{\textbf{Qualitative comparison of SD3}. Generated image examples before and after applying \method{} are on CelebA (left) and LAION-COCO (right).}
    \label{fig:qualitative_sd3}
\end{figure}

In Figure~\ref{fig:qualitative_sd3} and Figure~\ref{fig:qualitative_sd35}, we present qualitative examples of SD3 and SD3.5~\cite{Esser2024ScalingRF} before and after applying \method{} on CelebA and LAION-COCO. Across all resolutions, \method{} improves the overall image quality, providing better structure and textures of the generated images. In Figure~\ref{fig:qualitative_flux}, we show how \method{} improves the Flux-Dev model down to 64$\times$64 resolutions.
Overall, we observe the quality improvement and fewer artifacts across the resolutions. 
\textit{More qualitative examples are provided in the supplementary material.}

\section{Conclusion}
\label{sec:conclusion}

We present \method{}, a training-free test-time calibration method for direct low-resolution sampling in modern diffusion and flow-matching models.
Rather than modifying the sampling schedule or retraining the model, \method{} keeps the scheduled noise sequence fixed and recalibrates only the denoiser's noise-level conditioning, providing a lightweight and minimal adaptation mechanism for cross-resolution sampling.
Across multiple recent text-to-image models, we showed that this simple intervention consistently improves low-resolution generation quality while remaining compatible with both fixed and resolution-aware schedulers.
These results suggest that conditioning recalibration offers a practical and complementary tool for improving resolution flexibility in pretrained diffusion models.

\vspace{-0.1in}
\paragraph{Limitations and Future Work.}
While \method{} enhances cross-resolution performance, it does not fully solve the challenge of low-resolution generalization. Future work may explore integrating \method{} with learned adapters, dynamic token routing, or resolution-specific fine-tuning strategies.

\vspace{-0.1in}
\paragraph{Broader Impact.}
By enhancing sample quality at low resolutions, \method{} increases the adaptability of text-to-image systems to diverse deployment settings, including mobile and assistive applications. However, the ability to generate increasingly high-quality synthetic content may also exacerbate issues around misinformation and image provenance, highlighting the need for responsible use and effective detection mechanisms.

\bibliography{main}
\bibliographystyle{tmlr}
\end{document}